\def\year{2021}\relax
\documentclass[letterpaper]{article} 
\usepackage{aaai21}  
\usepackage{times}  
\usepackage{helvet} 
\usepackage{courier}  
\usepackage[hyphens]{url}  
\usepackage{graphicx} 
\usepackage{natbib}  
\usepackage{caption} 
\usepackage{amsthm}
\usepackage{amsmath}
\usepackage{amssymb}
\usepackage[ruled,vlined]{algorithm2e}
\usepackage{tikz}
\usepackage{pgfplots}
\usepackage{multirow}
\usepackage{booktabs}
\usepackage{natbib}
\usetikzlibrary{matrix}
\usetikzlibrary{calc}

\usepackage[textsize=tiny,
	disable,
	]{todonotes}
\usepackage{mathtools}
\usepackage{comment}

\newcommand{\frm}[2][]{\todo[color=red!60,linecolor={red!100},#1,size=\tiny]{#2}}

\newcommand{\lras}[2][]{\todo[color={violet!20},linecolor={violet!100},#1,size=\tiny]{#2}}

\newcommand{\rg}{RG}

\newcommand{\pom}{POM}
\newcommand{\pomA}{POM-10\%}

\newcommand{\pomC}{POM-30\%}

\newcommand{\dhc}{\ensuremath{\Gamma^{\text{LP}}}}
\newcommand{\dhcu}{\ensuremath{\Gamma^{\uncertainty}}}
\newcommand{\dhcf}{\ensuremath{\Gamma^{\text{\unreliability}}}}

\DeclareMathOperator{\cost}{cost}
\DeclareMathOperator{\pre}{pre}
\DeclareMathOperator{\post}{post}
\DeclareMathOperator{\vars}{vars}
\newtheorem{definition}{Definition}

\providecommand\tuple[1]{\ensuremath\langle#1\rangle}
\providecommand\Z{\ensuremath{\mathbb{Z}}}

\providecommand\Y[1]{\ensuremath{\mathsf{Y}_{#1}}}

\DeclareMathOperator{\variables}{\mathcal{V}}
\DeclareMathOperator{\operators}{\mathcal{O}}
\providecommand\domain[1]{\ensuremath{D({#1})}}

\DeclareMathOperator{\planningtask}{\Pi}
\DeclareMathOperator{\goalconditions}{\Gamma}
\DeclareMathOperator{\plan}{\pi}

\DeclareMathOperator{\occur}{\mathit{occur}}
\providecommand\initialstate{\ensuremath{s_{0}}}
\providecommand\goalstate{\ensuremath{s^{*}}}

\providecommand\h{\ensuremath{h}}
\providecommand\hoptimal{\ensuremath{h^{*}}}
\providecommand\hip{\ensuremath{h}^{\textup{IP}}}

\DeclareMathOperator{\grsolution}{\Gamma^{*}}
\DeclareMathOperator{\observations}{\Omega}

\providecommand\varvalue[2]{\ensuremath{#1[#2]}}
\providecommand\obs[1]{\ensuremath{\vec{#1}}}

\providecommand\constraints{\ensuremath{C}}
\providecommand\uncertainty{\ensuremath{\mu}}
\providecommand\unreliability{\ensuremath{\epsilon}}

\usepackage{subcaption}

\usepackage[switch]{lineno}

\title{An LP-Based Approach for Goal Recognition as Planning}
\author{
    Lu\'isa R. de A. Santos\textsuperscript{\rm 1}, 
    Felipe Meneguzzi\textsuperscript{\rm 2},
    Ramon Fraga Pereira\textsuperscript{\rm 3},
    Andr\'e G. Pereira\textsuperscript{\rm 1}
    \\
}
\affiliations{
    \textsuperscript{\rm 1}Federal University of Rio Grande do Sul, Brazil \\
    \textsuperscript{\rm 2}Pontifical  Catholic University of Rio Grande do Sul, Brazil \\
    \textsuperscript{\rm 3}Sapienza University of Rome, Italy \\
    
    \textsuperscript{\rm 1}\{lrasantos, agpereira\}@inf.ufrgs.br \\ 
    \textsuperscript{\rm 2}felipe.meneguzzi@pucrs.br \\
    \textsuperscript{\rm 3}pereira@diag.uniroma1.it\\ 

}
\pgfplotsset{compat=1.16}
\begin{document}

\maketitle

\begin{abstract}
\emph{Goal recognition} aims to recognize the set of candidate goals that are compatible with the observed behavior of an agent.
In this paper, we develop a method based on the \emph{operator-counting framework} that efficiently computes solutions that satisfy the observations and uses the information generated to solve goal recognition tasks. 
Our method reasons explicitly about both partial and noisy observations: estimating uncertainty for the former, and satisfying observations given the unreliability of the sensor for the latter. 
We evaluate our approach empirically over a large data set, analyzing its components on how each can impact the quality of the solutions. 
In general, our approach is superior to previous methods in terms of agreement ratio, accuracy, and spread. 
Finally, our approach paves the way for new research on \emph{combinatorial optimization} to solve goal recognition tasks.

\end{abstract}

\section{Introduction}

\emph{Goal recognition} as \emph{planning} consists of inferring the set of compatible goals from a set of goal candidates, given a planning task without a goal, and a sequence of observations. 
A solution for a goal recognition task is a subset of goal candidates that are compatible with the sequence of observations. 
A plan for the planning task with the reference goal, part of the set of goal candidates, generates the sequence of observations. This sequence may be partial, containing any number of observations from the plan.
Existing methods on goal recognition try to cope with three main classes of observation sequences: optimal~\cite{ramirez2009plan}, sub-optimal~\cite{ramirez2010probabilistic}, and noisy ~\cite{sohrabi2016plan}.
Since approaches to goal recognition as planning often employ standard planning technology to solve goal recognition tasks, many of them can benefit from improvements in the underlying planning technology~\cite{ramirez2009plan,martin2015fast,pereira2017landmark,HarmanSAAAI20}.\frm{Changed here, makes the text less staccato} 


%


Recent developments in planning include heuristics based on the \emph{operator-counting framework}, which combines the information of admissible heuristic functions through an \emph{integer program}~(IP)~\cite{pommerening2014lp}. 
These heuristics provide constraints that must be satisfied by every plan of the planning task. 
In general, the objective value of the \emph{linear program}~(LP), a linear relaxation of the integer program, is used as heuristic function to guide the search.
A major advantage of this framework is that it enables to reason and to manipulate the information of the heuristics directly. 

We develop an LP-based approach to solve goal recognition tasks, including five main contributions. 
First, we modify the operator-counting framework to efficiently compute solutions that satisfy the counts of observations of a goal recognition task. 
We also use this framework to estimate the cost of an optimal plan for each goal candidate in the task. Then, we use the information generated to solve the goal recognition task. 
Second, we show how to contrast the objective value of the modified linear program and the length of the sequence of observations to estimate the uncertainty of the decision of our approach which we use to improve our solution.
Third, we develop an approach to explicitly address noisy observations. 
Given the unreliability of the sensor of observations, we create an integer program that aims to automatically ignore noisy observations when computing solutions. 
Fourth, we show that higher heuristic values from lower bound heuristics for the reference goal predict the quality of our solution.
Finally, we modify the previous benchmarks to compare goal recognition methods by agreement ratio, showing that ours overcomes the state of the art. 



\section{Planning Task and Operator-Counting Framework}

\begin{figure}[t]
\centering
\scalebox{1}{
\includegraphics[]{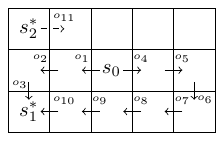}
}
\caption{A goal recognition task example.}
\label{fig:ex1}
\end{figure}

An \emph{SAS$^+$ planning task} is a tuple~${\planningtask = \tuple{\variables, \operators, \initialstate, \goalstate, \cost}}$, where $\variables$ is a set of \emph{variables}, $\operators$ is a set of \emph{operators}, $\initialstate$ is an \emph{initial state}, $\goalstate$ is a \emph{goal} condition, and $\cost$ a cost function. 
Each variable $v \in \variables$ has a finite domain $\domain{v}$. 
A \emph{state} is a complete assignment, a \emph{partial state} is a partial assignment of the variables over~$\variables$, $\vars(s)$ is the set of variables in a (partial) state~$s$, and $\varvalue{s}{v}$ is the value of variable $v$ in a (partial) state~$s$. 
The initial state~$\initialstate$ is a state, and the goal condition~$\goalstate$ is a partial state. 
A state~$s$ is consistent with a (partial) state~$s'$ if $s[v]=s'[v]$ for all $v\in\vars(s')$. 
Each operator~$o \in \operators$ is pair of partial states $\tuple{\pre(o),\post(o)}$ and an operator~$o$ is applicable to a state~$s$ if~$s$ is consistent with~$\pre(o)$. 
Applying~$o$ to~$s$ generates a new state~$s'$ such that~$s'[v]=\post(o)[v]$ for  $\vars(\post(o))$ and for the remaining variables $s'[v]=s[v]$. 
Function $\cost: \operators \rightarrow \Z^{+}_0$ assigns a non-negative cost to each operator $o\in \operators$ -- in this paper all operators have unit cost.
An $s$-plan~$\pi$ is a sequence of operators $\tuple{o_1,\ldots,o_n}$ such that there exists a sequence of states $\tuple{s_1=s, \ldots, s_{n+1}}$ where $o_i$ is applicable to $s_{i}$ and produces state $s_{i+1}$, and $s_{n+1}$ is consistent with $\goalstate$. 
The cost of a $s$-plan~$\pi$ is defined as $\cost(\pi) = \sum_{o \in \pi} \cost(o)$. 
An $\initialstate$-plan or a \emph{plan} is a solution to a planning task and is \emph{optimal} if its cost is minimal. 
Figure~\ref{fig:ex1} illustrates our running example where the agent performs cardinal movements and starts at~$\initialstate$. An optimal plan that reaches~$\goalstate_1$ for this task is $\pi=\tuple{o_1,o_2,o_3}$.

\begin{definition}[\bf Operator-Counting Constraint] Let $\planningtask$ be a planning task with operators $\operators$, and let $s$ be one of its states. 
	Let $\mathcal{Y}$ be a set of real-valued and integer variables, including an operator-counting non-negative integer variable $\Y{o}$ for each operator $o \in \operators$.
A set of linear inequalities over $\mathcal{Y}$ is an \emph{operator-counting constraint} for $s$ if for every valid $s$-plan~$\pi$, there exists a solution with $\Y{o} = \occur_{\pi}(o)$ for all $o\in\operators$ -- where $\occur_{\pi}(o)$ is the number of occurrences of operator~$o$ in the $s$-plan~$\pi$.
\end{definition}

In the example $\mathsf{Y_{o_3}+Y_{o_{10}}} \geq 1$ is an operator-counting constraint (and a landmark constraint) for goal~$\goalstate_1$ because the agent must use one of these operators to reach~$\goalstate_1$.

\begin{definition}[\bf Operator-Counting $\textup{IP}$/$\textup{LP}$ Heuristic]
The \emph{operator-counting integer program} $\textup{IP}^C$ for a set of operator-counting constraints $C$ for state $s$ is
\begin{align*}
    & \text{minimize} \sum_{o\in O} \cost(o)\Y{o}\\
    & \text{subject to $C$},\\
    & \Y{o}\in\Z^{+}_0.
\end{align*}
The $\textup{IP}$ \emph{heuristic} $h^{\textup{IP}}$ is the objective value of $\textup{IP}^C$, and the $\textup{LP}$ \emph{heuristic} $h$ is the objective value of its linear relaxation. If the $\textup{IP}$ or $\textup{LP}$ is infeasible, the heuristic estimate is $\infty$.
\end{definition} 


\section{Goal Recognition as Planning}

We formally define the task of \emph{goal recognition as planing} as a tuple $\langle \planningtask_{\textup{P}}, \goalconditions, \observations\rangle$, where $\planningtask_{\textup{P}}$ is a planning task without a goal condition, $\goalconditions$ is a set of goal candidates, and $\observations$ is a sequence of observations. 
Observation $\obs{o}$ corresponds to operator $o$. 
For readability, we abuse notation and equate operators to observations throughout the paper when convenient. 


\begin{definition}[\bf Observation Compliance]
Let~$\plan=\tuple{o_1, \ldots, o_n}$ be a plan for a planning task~$\planningtask$ and $\observations=\tuple{\obs{o}_1, \ldots, \obs{o}_m}$ a sequence of observations.
Plan $\pi$ \emph{complies} with $\observations$ if a monotonic function $f : [1,m] \mapsto [1,n]$ exists that maps all operator indexes in $\observations$ to indexes in $\plan$, such that $ \obs{o}_i = o_{f(i)}$. 
\end{definition}

We define three classes of sequences of observations: \textit{optimal} and \textit{sub-optimal} (Definition~\ref{def:observations_opt_subopt}) observations, and \textit{noisy optimal/sub-optimal} observations (Definition~\ref{def:observations_noisy}).

\begin{definition}[\bf Sequence of Observations] \label{def:observations_opt_subopt}
Let~$\pi=\tuple{o_1,\ldots,o_n}$ be a plan for the planning task~$\planningtask$.
Then, a \emph{sequence of observations}~$\observations$ is a sequence of operators extracted from the plan~$\pi$ maintaining their relative order. The sequence may be partial, containing any number of operators from the plan~$\pi$.
An \emph{optimal} sequence of observations is extracted from an optimal plan and a \emph{sub-optimal} sequence of observations is extracted from a sub-optimal plan. An optimal/sub-optimal observation is part of an optimal/sub-optimal sequence of observations. 
\end{definition}

\begin{definition}[\bf Noisy Observations]
\label{def:observations_noisy}


A \emph{noisy} sequence of observations~$\observations$ is a sequence observations extracted from~$\plan$ that is augmented with at least one observation~$\operators - \plan$, which is inserted in any position of the extracted sequence. 


\end{definition}

We extend the standard definition from \citeauthor{ramirez2009plan}~\shortcite{ramirez2009plan} of an \emph{exact solution set} for a goal recognition task to also consider sub-optimal observation sequences (Definition~\ref{def:gr_solutionset}) and call it \textit{reference solution set}. 
We define the \emph{reference solution set} as a subset of the goal candidates such that there exists a complying plan as sub-optimal as or less than the plan that generated the observations for the reference goal. 


\begin{definition}[\bf Reference Solution Set]\lras{changed here}
\label{def:gr_solutionset}
Let~$\langle \planningtask_{\textup{P}}, \goalconditions, \observations\rangle$ be a \emph{goal recognition task} and $\planningtask$ a planning task with the goal condition  $\goalstate\in\goalconditions$ (the reference goal).
Let~$\pi^*$ be an optimal plan for~$\planningtask$, and let~$\plan$ be a plan for~$\planningtask$ from which $\observations$ is extracted. 
Let~$h^*_{\observations}(s_0,\goalstate_{i})$ be the cost of an optimal plan for~$\planningtask$ restricted to the set of plans that comply with~$\observations$, and~$h^*(s_0,\goalstate_{i})$ be the cost for an optimal plan for~$\planningtask$, both with $\goalstate_{i}\in\Gamma$. 
$h^*_{\observations}(s_0,\goalstate_{i})$ and $h^*(s_0,\goalstate_{i})$ are equal to $\infty$ if no plan exists.
Then, the \lras{changed here}\emph{reference solution set} for the \emph{goal recognition task} is
	$$\grsolution = \{\goalstate_{i} \in \goalconditions \mid   
	   \frac{h^*_{\observations}(s_0,\goalstate_{i})}{h^*(s_0,\goalstate_{i})} \leq \frac{\cost(\plan)}{\cost(\pi^*)} \land h^*_{\observations}(s_0,\goalstate_{i})\neq\infty \}$$
\end{definition}

In Figure~\ref{fig:ex1} we show a goal recognition task
with goal candidates $\goalconditions = \{\goalstate_{1}, \goalstate_{2}\}$.
Suppose that~$\goalstate_1$ is the reference goal. Then, $\observations_1 = \tuple{ \obs{o}_1}$ is an optimal sequence of observations because it is extracted from the optimal plan~$\pi_1=\tuple{o_1,o_2,o_3}$, $\observations_2 = \tuple{ \obs{o}_5,  \obs{o}_7,  \obs{o}_9}$ and $\observations_3 = \tuple{ \obs{o}_4,  \ldots,  \obs{o}_{10}}$ are sub-optimal sequences of observations because they are extracted from the sub-optimal plan~$\pi_2=\tuple{o_4,\ldots,o_{10}}$, and $\observations_4 = \tuple{ \obs{o}_4,\ldots,\obs{o}_{10},  \obs{o}_{11}}$ is a sub-optimal and noisy sequence of observations because it was extracted from~$\pi_2$ and the observation of~$\obs{o}_{11}$ was added. 
The \lras{changed here}reference solution set for goal recognition tasks with noisy observations is computed ignoring noisy observations in the sequence of observations. 
The \lras{changed here}reference solution set for any of these observation sequences with respective plans is $\Gamma^*_i=\{\goalstate_1\}$. 
For example, $h^*_{\observations_4}(s_0,\goalstate_1)=7$,  $h^*_{\observations_4}(s_0,\goalstate_2)=9$, $\cost(\pi_2)/\cost(\pi^*)=7/3$
and thus $\Gamma^*_4=\{\goalstate_1\}$.

\section{LP-Based Goal Recognition}

We now develop an LP-based goal recognition method that expands the operator-counting framework with \emph{observation-counting constraints}. 

\subsubsection{Observation-Counting Constraints} 

We now introduce an IP/LP heuristic which expands the operator-counting framework with a set of \emph{observation-counting constraints}.  
Definition~\ref{def:complying-constraints} formally introduces the set of \emph{observation-counting constraints} and the integer program that ensures that the solution computed satisfies all observation counts.

\begin{definition}[\bf Satisfying IP/LP heuristic]
\label{def:complying-constraints}
Let~$\mathcal{Y}^{\observations}$ be a set of non-negative integer variables with a variable~$\Y{ \obs{o}}$ for each operator $o \in \operators$. 
Let $\occur_{\Omega}(o)$ be the number of occurrences of operator~$o$ in $\observations$. 
Then, the \emph{satisfying} integer program~$\textup{IP}^{\constraints}_{\observations}$ for a set of operator-counting constraints~$\constraints$, a set of \emph{observation-counting constraints}, and sequence of observations~$\observations$ for state $s$ is
\begin{align}
    & \text{minimize} \sum_{o\in O} \cost(o)\Y{o} & & &\nonumber\\
    & \text{subject to $C$}, & & &\nonumber\\    
    & \Y{ \obs{o}}\leq \occur_{\Omega}(o)                                 & \text{for all } o \in \operators \label{ceq1} \\
    & \Y{ \obs{o}}\leq \Y{o}                                                 & \text{for all } o \in \operators \label{ceq2} \\ 
    & \sum_{\Y{ \obs{o}} \in \mathcal{Y}^{\observations}} \Y{ \obs{o}} \geq |\observations|   & \label{ceq3}\\
    & \Y{o},\Y{ \obs{o}}\in\Z^{+}_0. \nonumber
\end{align}
The \emph{satisfying} $\textup{IP}$ \emph{heuristic} $\hip_{\observations}$ is the objective value of $\textup{IP}^{\constraints}_{\observations}$, and the \emph{satisfying} $\textup{LP}$ \emph{heuristic} $\h_{\observations}$ is the objective value of its linear relaxation. 
If the $\textup{IP}$ or $\textup{LP}$ is infeasible, the heuristic estimate is $\infty$.
\end{definition} 

In the integer program, the set of constrains~(\ref{ceq1}) limits the value of each~$\Y{ \obs{o}}$ by the number of occurrences of the operator~$o$ in $\observations$. 
Next, the set of constraints~(\ref{ceq2}) binds the two sets~$\Y{ \obs{o}}$ and~$\Y{o}$ of variables. 
This set of constraints guarantees that~$\Y{o}$ acts as an upper bound for~$\Y{ \obs{o}}$. 
Thus, to increase the count of~$\Y{ \obs{o}}$ the integer program must first increase the count of~$\Y{o}$ which is minimized in the objective function and restricted by the set of operator-counting constraints~$\constraints$. 
Finally, constraint~(\ref{ceq3}) ensures all observations are satisfied, since each~$\Y{ \obs{o}}$ is limited by the number of times~$o$ appears in $\observations$. 
While simpler models can compute the same objective value, we show how explicit information about the observations enables us to reason about noisy observations.

    
\frm{The note note note thing here is boring, I will leave it here, but we should edit this for the final version.}
Note that $\h_{\observations}(s,\goalstate{}) \leq \hoptimal_{\observations}(s,\goalstate{})$ for all states~$s$ of the planning task.
First note that $\h$ is \emph{admissible}~\cite{pommerening2014lp} and that a complying plan~$\pi$ can always satisfy~$\textup{IP}^C_{\observations}$. 
Note that the only difference between~$\textup{IP}^C_{\observations}$ and $\textup{IP}^C$ are the \emph{observation-counting constraints}. 
These constraints only restrict the set of plans that can satisfy~$\textup{IP}^C_{\observations}$ to the set of plans that satisfy all observations. 
If~$\pi$ is an optimal $s$-plan that complies with~$\Omega$ ($\h^*_{\observations}(s,\goalstate{})=\cost(\pi)$), then there is a solution for~$\textup{IP}^C_{\observations}/\textup{LP}^C_{\observations}$ where $\Y{ \obs{o}} = \Y{o} = \occur_{\pi}(o)$.

We use the~$\h_{\observations}$ heuristic to estimate a lower bound on the cost of an optimal plan that satisfies all observations in~$\observations$ for each goal candidate in~$\goalconditions$. 
However, this information is insufficient to estimate the solution set because the goal candidate with the least~$\h_{\observations}$-value is not necessarily the most likely one. 
Consider a goal recognition task with two goal candidates. 
The first goal candidate has an optimal cost plan that can be extended with one operator to satisfy the single observation in~$\observations$. 
The second goal candidate has an optimal cost plan that complies with the observation in~$\observations$. 
In this example the plan for the first goal costs less than the plan for the second goal.
In this example only the first goal candidate would be included in the solution set. However, we argue that the second goal is more likely to be part of the \lras{changed here}reference solution set since it is the only goal candidate with a complying optimal plan. 
Therefore, we normalize the values of~$h_{\observations}$ with estimates of the costs of the original optimal solution -- without satisfying the observations. 
Like previous methods, the idea is to select the goals that have plans that satisfy all observations with the least additional cost. 
For this, we use the value of the original operator-counting heuristic~$\h$. 
Having~$\h_{\observations}$ and~$\h$ for each goal candidate we can compute the following solution set: 
\begin{equation}
\delta_{\text{min}} = \min_{\goalstate_i\in\goalconditions\text{ }:\text{ }\h_{\observations}(\initialstate,\goalstate_i) < \infty} \{\h_{\observations}(\initialstate,\goalstate_i) - \h(\initialstate,\goalstate_i)\}
\label{equation-delta}
\end{equation}
\begin{equation}
\dhc = \{\goalstate_i \in \goalconditions \mid \h_{\observations}(\initialstate,\goalstate_{i}) - \h(\initialstate,\goalstate_{i}) = \delta_{\text{min}}\}
\label{equation-solution-set}
\end{equation}

Equation~\ref{equation-delta} computes the minimum difference~$\delta_{\text{min}}$ between the lower bound cost of an optimal plan that satisfies observations and the lower bound cost of an optimal plan (ignoring observations). The ~$\delta_{\text{min}}$ value only considers goal candidates with bounded estimates for plans that satisfy observations ($\h_{\observations}(\initialstate,\goalstate) < \infty$).
Equation~\ref{equation-solution-set} computes the solution set~$\Gamma^\textup{LP}$ by selecting all goals with a difference between the estimates equal to $\delta_{\text{min}}$. Note that~$\Gamma^\textup{LP}$ is an approximated solution and not equal to the \lras{changed here}reference solution set~$\grsolution$.
In our running example, consider a goal recognition task with $\observations = \tuple{ \obs{o}_5,  \obs{o}_7, \obs{o}_9}$. 
Then cost of $h_{\observations}(\initialstate,\goalstate_1)$ and $h(\initialstate,\goalstate_1)$ are $7$ and $3$, and costs of $h_{\observations}(\initialstate,\goalstate_2)$ and $h(\initialstate,\goalstate_2)$ are $9$ and $3$. 
Thus, $\delta_{\text{min}}$ equals to $4$, and we return $\Gamma^{\textup{LP}} = \{\goalstate_1\}$.

\subsubsection{Addressing Noisy Observations}

In most realistic settings, unreliable sensors may add noisy observations to the sequence of observations. 
Consider a goal recognition task in our running example with $\observations = \tuple{\obs{o}_4,\ldots,\obs{o}_{10},\obs{o}_{11}}$. 
Then, $\h_{\observations}(\initialstate,\goalstate_1)= 13$ and $\h_{\observations}(\initialstate,\goalstate_2)= 11$. 
In this situation we would have $\delta_{\text{min}} = 8$, and  $\Gamma^\textup{LP} = \{\goalstate_2\}$. However, the observation~$~\obs{o}_{11}$ is unlikely to be part of any plan that generates the sequence of observations for either of the two goals. 
Evaluating precisely which observations are unlikely to be part of plans for a goal is a hard 
problem that requires solving a planning task multiple times, or, as \citeauthor{sohrabi2016plan}~\shortcite{sohrabi2016plan} do, generating multiple plans.  
In spite of that, we can the estimate the solution for this problem in polynomial time using the linear relaxation of an integer program. 
Specifically, we modify the integer program to try to automatically identify noisy observations given the unreliability of the sensors. 
The main modification is to replace constraint~(\ref{ceq3}) in the integer program~$\textup{IP}_C^{\observations}$ with constraint~(\ref{eq-noisy}). 
We call the solution set using this heuristic~$\dhcf$. 

\begin{equation}
\sum_{\Y{ \obs{o}} \in \mathcal{Y}^{\observations}} \Y{ \obs{o}} \geq |\observations| -  \lfloor|\observations| * \unreliability\rfloor
\label{eq-noisy}
\end{equation}

\noindent where $\unreliability$ is the \emph{unreliability rating} of the sensor that represents the expected percentage of mistaken observations. 
This new constraint requires that at least $|\observations| -  \lfloor|\observations| * \unreliability\rfloor$ observations be satisfied by the solution found. 
\frm{A note here. It should be either ``This new constraint requires \textbf{that} at least $|\observations| -  \lfloor|\observations| * \unreliability\rfloor$ observations \textbf{be} satisfied by the solution found. '' \textbf{or} ``This new constraint requires at least $|\observations| -  \lfloor|\observations| * \unreliability\rfloor$ observations \textbf{to be} satisfied by the solution found. ''}
If $\unreliability = 0$, all observations must be satisfied. 
If $0 < \unreliability < 1$, some observations can be automatically ignored in order to minimize the objective value of~$h_{\observations}$ for each goal candidate. 
Consider our running example with $\observations = \tuple{ o_4,\ldots,o_{10},o_{11}}$ and $\unreliability = 0.2$. 
Then, the integer program~$\textup{IP}^C_{\observations}$ has to satisfy~$7$ observations. 
Then, $h_{\observations}(\initialstate,\goalstate_1)= 7$ and $h_{\observations}(\initialstate,\goalstate_2)= 9$. In this situation we would have $\delta_{\text{min}} = 4$, and~$\dhcf = \{\goalstate_1\}$.

\subsubsection{Measuring Uncertainty}

The main idea of this approach is that if~$\observations$ has a lower percentage of observations we should be more careful in our decision. However, if the percentage is higher we can be more confident. 
A goal recognition task does not provide the percentage of observations in~$\observations$. 
However, we can estimate this information using our heuristic~$\h_{\observations}$.
Since $\h_{\observations}$ provides a lower bound on the cost of a plan that satisfies the observations 
the difference (if any) between $\h_{\observations}$ and $|\observations|$ must be due to missing observations. 
Note that under $100\%$ observability $\h_{\observations} = |\observations|$, 
and with lower degrees of observability~$\h_{\observations}$ may select operators that are not in~$\observations$ to satisfy the operator-counting constraints~$\constraints$. 
Thus, in lower degrees of observability the difference between $\h_{\observations}$ and $|\observations|$ is likely to increase. 
This information allows us to estimate the degree of observability as follows:
%
\begin{equation}
\uncertainty = 1 + \dfrac{\max_{\goalstate_i\in\Gamma^{\textup{LP}}}\{h_{\observations}(\initialstate,\goalstate_i)\} - |\observations|}{\max_{\goalstate_i\in\Gamma^{\textup{LP}}}\{h_{\observations}(\initialstate,\goalstate_i)\}}
\label{eq-uncertainty}
\end{equation}

where $\uncertainty$ is the \textit{uncertainty ratio}. 
This value is computed by first selecting the goal candidates using Equation~\ref{equation-solution-set}, and then selecting the goal candidate in $\dhc$ with maximum $\h_{\observations}$. 
Having~$\uncertainty$ we can compute the new solution set~$\dhcu$ that considers uncertainty:
\begin{equation}
\dhcu = \{\goalstate_i \in \goalconditions \mid \h_{\observations}(\initialstate,\goalstate_i) - \h(\initialstate,\goalstate_i) \leq \delta_{\text{min}} *\uncertainty \}
\label{equation-solution-set-uncertainty}
\end{equation}

Consider our running example with $\observations = \tuple{ \obs{o}_6}$.
Then, $\h_{\observations}(\initialstate,\goalstate_1)= 7$ and $\h_{\observations}(\initialstate,\goalstate_2)= 9$. In this situation we would have $\delta_{\text{min}} = 4$, and  $\dhc = \{\goalstate_1\}$. However, we would argue that having only one observation is insufficient to make a precise decision. Using uncertainty we would have  
$\uncertainty=1+6/7$ and $\dhcu=\{\goalstate_1, \goalstate_2\}$.

\section{Experimental Results}

We conducted extensive empirical experiments to show the effectiveness of our methods in three ways. 
First, we evaluate how each source of operator-counting constraints impacts the quality of our solutions. 
Second, we assess the performance of our methods that explicitly address low observability and noise. 
Finally, we compare our approach with previous ones. 
We ran all experiments with Ubuntu running over an Intel Core i7 930 CPU ($2.80$\,GHz) with a memory limit of $1$\,GB, all methods solved each goal recognition task under a time limit of five seconds. 
Our implementation uses Fast Downward version 19.06~\cite{helmert2006jair}, a Python prepossessing layer, and the CPLEX 12.10 LP solver.\footnote{Source-code and benchmark are available at: \texttt{https://bit.ly/lp-goal-recognition}} 


\subsubsection{Benchmark Domains and Data sets}

We create a new benchmark by adapting the one introduced by~\citeauthor{pereira2017landmark}~\shortcite{pereira2017landmark} to use the \emph{agreement ratio} evaluation metric from~\citeauthor{ramirez2009plan}~\shortcite{ramirez2009plan}. 
For each domain we create three planning tasks each (except for \textsc{IPC-Grid}, in which we create four) with four reference goals each that we use to compute the plans from which we extract the sequence of observations. 
We compute optimal and sub-optimal plans for each pair of planning task and reference goal creating two data sets. 
To compute sub-optimal plans we use weighted $A^*$ with $w = 2$~\cite{pohl1970heuristic}. 

Following previous work, we experiment with five different levels of observability: 10\%, 30\%, 50\%, 70\% and 100\%. 
We only generate one sequence of observations for 100\% of observability, and three different random observation sequences from the same plan for other percentages, generating 208 goal recognition tasks in total for \textsc{IPC-Grid} and 156 for each of the other domains in each data set (optimal and sub-optimal). 
For each data set we also create a corresponding noisy data set by adding $\lceil|\observations| * 0.2\rceil$ randomly generated observations in each sequence of observations--- i.e. the fault rate of the sensor is 20\%. Three different noisy sequences are generated for each original sequence.
For each goal recognition task we add at least five randomly generated candidate goal conditions. 
In total we have $8,288$ goal recognition tasks divided in four data sets.
In order to create the new benchmarks, we compute the \lras{changed here}reference solution set~$\grsolution$ for each goal recognition task for optimal and sub-optimal data sets. 
Thus, for each goal candidate of each base task we solve a planning task twice.

We evaluate the methods using three metrics: \emph{agreement ratio}, \emph{accuracy} and \emph{spread}. 
The \emph{agreement ratio} is defined as the intersection over union~$|\grsolution \cap \goalconditions|/|\grsolution \cup \goalconditions|$ of the \lras{changed here}reference solution set~$\grsolution$ against the solution~$\goalconditions$ provided by the method. 
The \emph{accuracy} is $1$ if the solution set chosen by the evaluated method contains the reference goal and $0$ otherwise. Note that we use a slight modified of accuracy in other to compare to~\cite{pereira2017landmark}.
The \emph{spread} is the size of the solution set chosen by the evaluated method.
Table~\ref{benchmark-domains} summarizes the information about the data sets. 
The domains we use are \textsc{Blocks World}, \textsc{Depots}, \textsc{Driverlog}, \textsc{DWR}, \textsc{IPC Grid}, \textsc{Ferry}, \textsc{Logistics}, \textsc{Miconic}, \textsc{Rovers}, \textsc{Satellite}, \textsc{Sokoban} and \textsc{Zeno Travel}. 
Due to space restrictions, we summarize results for all domains as averages in \textsc{Other}, except for \textsc{Blocks World}, \textsc{IPC Grid} and \textsc{Sokoban}.
For each domain row, $|\goalconditions|$ is the average number of candidate goals.
Columns $|\observations|$ and $|\grsolution|$ show the average sizes of the observations and the \lras{changed here}reference solution set, respectively. 
The average size of the plan with 100\% of observability indicates the size of the plan computed for the reference goal.
As expected, the average sizes $|\observations|$ and $|\grsolution|$ are larger for the sub-optimal data set than for the optimal data set. 

\begin{table}[htb]
\centering
\fontsize{9.}{9.}\selectfont 
\setlength\tabcolsep{1.8pt}
\begin{tabular}{cc|c|cc|cc}
\toprule
\multicolumn{3}{c}{} & \multicolumn{2}{c}{Optimal} & \multicolumn{2}{c}{Sub-Optimal}\\
\midrule
\# & \% & $|\goalconditions|$ & $|\observations|$ & $|\grsolution|$ & $|\observations|$ & $|\grsolution|$ \\
\midrule
\multirow{5}{*}{ \rotatebox[origin=c]{90}{\textsc{blocks}}}%
	 & 10
 & \multirow{5}{*}{20.33} & 1.25 & 8.0 & 1.42 & 7.61\\
	 & 30
 &  & 3.08 & 3.97 & 3.83 & 3.58\\
	 & 50
 &  & 4.42 & 2.5 & 5.92 & 3.19\\
	 & 70
 &  & 6.67 & 1.94 & 8.5 & 2.53\\
	 & 100
 &  & 8.83 & 1.83 & 11.83 & 2.25\\\hline
 \multirow{5}{*}{ \rotatebox[origin=c]{90}{\textsc{ipc-grid}}}%
	 & 10
 & \multirow{5}{*}{7.5} & 1.63 & 2.71 & 2.06 & 1.58\\
	 & 30
 &  & 4.0 & 1.21 & 5.56 & 1.4\\
	 & 50
 &  & 6.19 & 1.13 & 8.88 & 1.35\\
	 & 70
 &  & 8.69 & 1.04 & 12.56 & 1.31\\
	 & 100
 &  & 11.88 & 1.0 & 17.25 & 1.5\\\hline
 \multirow{5}{*}{ \rotatebox[origin=c]{90}{\textsc{sokoban}}}%
	 & 10
 & \multirow{5}{*}{8.67} & 2.33 & 2.11 & 3.33 & 1.83\\
	 & 30
 &  & 6.5 & 1.25 & 8.67 & 1.28\\
	 & 50
 &  & 10.33 & 1.22 & 13.75 & 1.33\\
	 & 70
 &  & 14.67 & 1.03 & 19.33 & 1.36\\
	 & 100
 &  & 20.17 & 1.0 & 27.0 & 1.33\\\hline
\multirow{5}{*}{ \rotatebox[origin=c]{90}{\textsc{Other}}}%
	 & 10
 & \multirow{5}{*}{6.89} & 1.85 & 3.01 & 2.46 & 2.32\\
	 & 30
 &  & 4.69 & 1.61 & 6.37 & 1.45\\
	 & 50
 &  & 7.52 & 1.21 & 10.04 & 1.21\\
	 & 70
 &  & 10.61 & 1.1 & 14.13 & 1.15\\
	 & 100
 &  & 14.51 & 1.06 & 19.55 & 1.08\\
\bottomrule
\end{tabular}\\
\caption{Key properties of each experimental domain.}
\label{benchmark-domains} 
\end{table}

\begin{table}
\centering
\fontsize{9.}{9.}\selectfont
\setlength\tabcolsep{1.5pt}
    \addtolength{\leftskip} {-0.6cm} 
\begin{tabular}{cc|ccc|ccc||ccc|ccc}
\toprule
\multicolumn{2}{c}{} & \multicolumn{6}{c}{Optimal} & \multicolumn{6}{c}{Sub-Optimal}\\
\cmidrule(lr){3-8}\cmidrule(lr){9-14}
\# & \% & S & L & P & S, L & L, P & S, P%
& S & L & P & S, L & L, P & S, P\\
\midrule
\multirow{5}{*}{ \rotatebox[origin=c]{90}{\textsc{blocks}}}%
 & 10 & \textbf{0.45} & 0.42 & 0.44 & \textbf{0.45} & 0.41 & 0.44 & \textbf{0.44} & 0.41 & 0.39 & \textbf{0.44} & 0.39 & 0.41\\ & 30 & \textbf{0.43} & 0.33 & \textbf{0.43} & 0.43 & \textbf{0.47} & \textbf{0.47} & \textbf{0.5} & 0.44 & 0.41 & \textbf{0.5} & 0.44 & 0.49\\ & 50 & \textbf{0.55} & 0.46 & 0.44 & 0.55 & 0.58 & \textbf{0.59} & 0.5 & 0.37 & \textbf{0.51} & 0.5 & \textbf{0.57} & 0.55\\ & 70 & \textbf{0.75} & 0.54 & 0.58 & 0.75 & 0.81 & \textbf{0.85} & \textbf{0.64} & 0.45 & 0.55 & 0.64 & 0.69 & \textbf{0.71}\\ & 100 & \textbf{0.82} & 0.58 & 0.62 & 0.82 & 0.88 & \textbf{0.92} & \textbf{0.74} & 0.52 & 0.58 & 0.74 & 0.79 & \textbf{0.84}\\\hline
\multirow{5}{*}{ \rotatebox[origin=c]{90}{\textsc{ipc-grid}}}%
 & 10 & 0.65 & \textbf{0.92} & 0.4 & 0.87 & \textbf{0.92} & 0.68 & 0.6 & \textbf{0.86} & 0.25 & 0.76 & \textbf{0.86} & 0.63\\ & 30 & 0.73 & \textbf{0.97} & 0.25 & 0.93 & \textbf{0.97} & 0.78 & 0.69 & \textbf{0.88} & 0.23 & 0.82 & \textbf{0.88} & 0.71\\ & 50 & 0.83 & \textbf{0.97} & 0.27 & 0.96 & \textbf{0.97} & 0.9 & 0.81 & \textbf{0.89} & 0.29 & 0.84 & \textbf{0.89} & 0.87\\ & 70 & 0.9 & \textbf{0.97} & 0.3 & \textbf{0.97} & \textbf{0.97} & 0.95 & 0.87 & \textbf{0.91} & 0.08 & 0.89 & \textbf{0.91} & 0.89\\ & 100 & \textbf{1.0} & \textbf{1.0} & 0.23 & \textbf{1.0} & \textbf{1.0} & \textbf{1.0} & \textbf{0.94} & \textbf{0.94} & 0.05 & \textbf{0.94} & \textbf{0.94} & \textbf{0.94}\\\hline
\multirow{5}{*}{ \rotatebox[origin=c]{90}{\textsc{sokoban}}}%
 & 10 & \textbf{0.38} & \textbf{0.38} & 0.24 & \textbf{0.39} & 0.34 & 0.31 & \textbf{0.38} & 0.3 & 0.24 & \textbf{0.52} & 0.25 & 0.36\\ & 30 & \textbf{0.59} & 0.41 & 0.14 & \textbf{0.75} & 0.38 & 0.59 & \textbf{0.72} & 0.43 & 0.14 & \textbf{0.77} & 0.37 & 0.68\\ & 50 & \textbf{0.82} & 0.53 & 0.21 & \textbf{0.92} & 0.49 & 0.82 & \textbf{0.77} & 0.51 & 0.17 & \textbf{0.79} & 0.41 & \textbf{0.79}\\ & 70 & \textbf{0.93} & 0.73 & 0.21 & \textbf{0.99} & 0.62 & 0.93 & \textbf{0.85} & 0.58 & 0.17 & 0.8 & 0.51 & \textbf{0.85}\\ & 100 & \textbf{0.96} & 0.85 & 0.23 & \textbf{1.0} & 0.81 & 0.96 & \textbf{0.88} & 0.73 & 0.22 & 0.83 & 0.72 & \textbf{0.88}\\\hline%
\multirow{5}{*}{ \rotatebox[origin=c]{90}{\textsc{Other}}}%
 & 10 & 0.71 & \textbf{0.73} & 0.63 & \textbf{0.78} & 0.72 & 0.69 & \textbf{0.63} & \textbf{0.63} & 0.55 & \textbf{0.72} & 0.63 & 0.64\\ & 30 & \textbf{0.71} & \textbf{0.71} & 0.54 & \textbf{0.82} & 0.7 & 0.73 & 0.67 & \textbf{0.7} & 0.54 & \textbf{0.78} & 0.69 & 0.69\\ & 50 & \textbf{0.81} & 0.78 & 0.58 & \textbf{0.88} & 0.77 & 0.82 & \textbf{0.8} & 0.78 & 0.61 & \textbf{0.87} & 0.78 & 0.83\\ & 70 & \textbf{0.91} & 0.87 & 0.63 & \textbf{0.96} & 0.87 & 0.92 & \textbf{0.9} & 0.87 & 0.64 & \textbf{0.94} & 0.88 & 0.9\\ & 100 & \textbf{0.96} & 0.94 & 0.66 & \textbf{0.98} & 0.95 & 0.95 & \textbf{0.95} & 0.93 & 0.66 & \textbf{0.97} & 0.95 & 0.95\\\midrule
\multicolumn{2}{c|}{AVG}  & \textbf{0.79} & 0.77 & 0.54 & \textbf{0.86} & 0.78 & 0.8 & \textbf{0.76} & 0.74 & 0.52 & \textbf{0.82} & 0.75 & 0.78\\
\bottomrule
\end{tabular}\\
\caption{Agreement ratio for constraint sets state equation~$h^{\text{SEQ}}_{\observations}$ (S), landmarks~$h^{\text{LMC}}_{\observations}$ (L), and post-hoc~$h^{\text{PhO}}_{\observations}$ (P).}
\label{agr-constraints}
\end{table}

\begin{table*}
\centering
\fontsize{9}{9.}\selectfont
\setlength\tabcolsep{2pt}
\begin{tabular}{cc|ccc|ccc|ccc|ccc|ccc|ccc}
\toprule
 \multicolumn{2}{c}{} & \multicolumn{18}{c}{Optimal}\\
\cmidrule(lr){3-20}
 \multicolumn{2}{c}{} & \multicolumn{3}{c}{\dhc} & \multicolumn{3}{c}{\dhcu} & \multicolumn{3}{c}{\rg} & \multicolumn{3}{c}{\pom} & \multicolumn{3}{c}{\pomA} & \multicolumn{3}{c}{\pomC}\\%
\cmidrule(lr){3-5}\cmidrule(lr){6-8}\cmidrule(lr){9-11}\cmidrule(lr){12-14}\cmidrule(lr){15-17}\cmidrule(lr){18-20}\\
\# & \% %
& AGR & ACC & SPR & AGR & ACC & SPR & AGR & ACC & SPR & AGR & ACC & SPR & AGR & ACC & SPR & AGR & ACC & SPR\\
\midrule
\multirow{5}{*}{ \rotatebox[origin=c]{90}{\textsc{blocks}}}%
 & 10 & 0.44 & 0.86 & 7.53 & 0.44 & 0.86 & 7.56 & \textbf{0.47} & 0.92 & 9.83 & 0.06 & 0.17 & 1.44 & 0.13 & 0.47 & 4.06 & 0.38 & 1.0 & 18.14\\ & 30 & \textbf{0.46} & 0.78 & 2.5 & 0.44 & 0.86 & 4.67 & 0.45 & 0.92 & 5.56 & 0.21 & 0.39 & 1.17 & 0.3 & 0.75 & 2.94 & 0.24 & 1.0 & 15.25\\ & 50 & 0.59 & 0.89 & 3.03 & 0.52 & 0.89 & 3.86 & \textbf{0.62} & 0.97 & 3.69 & 0.33 & 0.58 & 1.25 & 0.37 & 0.81 & 3.08 & 0.25 & 0.97 & 12.17\\ & 70 & \textbf{0.85} & 0.97 & 1.83 & 0.76 & 0.97 & 2.42 & 0.81 & 1.0 & 2.22 & 0.51 & 0.72 & 1.14 & 0.45 & 0.94 & 2.19 & 0.25 & 1.0 & 9.22\\ & 100 & \textbf{0.92} & 1.0 & 1.67 & \textbf{0.92} & 1.0 & 1.67 & 0.9 & 1.0 & 2.08 & 0.59 & 1.0 & 1.67 & 0.55 & 1.0 & 1.92 & 0.31 & 1.0 & 6.42\\\hline%
 \multirow{5}{*}{ \rotatebox[origin=c]{90}{\textsc{ipc-grid}}}%
 & 10 & 0.87 & 0.94 & 2.67 & 0.88 & 0.96 & 2.69 & \textbf{0.91} & 1.0 & 3.23 & 0.47 & 0.75 & 2.35 & 0.55 & 0.98 & 4.38 & 0.49 & 1.0 & 6.25\\ & 30 & 0.93 & 0.96 & 1.15 & 0.94 & 0.98 & 1.17 & \textbf{0.99} & 1.0 & 1.25 & 0.85 & 0.98 & 1.52 & 0.81 & 1.0 & 1.96 & 0.64 & 1.0 & 3.17\\ & 50 & 0.96 & 0.98 & 1.08 & 0.96 & 0.98 & 1.08 & \textbf{1.0} & 1.0 & 1.13 & 0.86 & 1.0 & 1.44 & 0.86 & 1.0 & 1.56 & 0.77 & 1.0 & 2.15\\ & 70 & 0.97 & 0.98 & 1.06 & 0.97 & 0.98 & 1.06 & \textbf{1.0} & 1.0 & 1.04 & 0.97 & 0.98 & 1.02 & 0.97 & 0.98 & 1.02 & 0.93 & 0.98 & 1.15\\ & 100 & \textbf{1.0} & 1.0 & 1.0 & \textbf{1.0} & 1.0 & 1.0 & \textbf{1.0} & 1.0 & 1.0 & \textbf{1.0} & 1.0 & 1.0 & \textbf{1.0} & 1.0 & 1.0 & \textbf{1.0} & 1.0 & 1.0\\\hline%
 \multirow{5}{*}{ \rotatebox[origin=c]{90}{\textsc{sokoban}}}%
 & 10 & 0.39 & 0.53 & 2.08 & 0.38 & 0.61 & 2.94 & \textbf{0.4} & 0.81 & 4.86 & 0.28 & 0.53 & 2.14 & 0.32 & 0.89 & 4.39 & 0.26 & 0.97 & 7.0\\ & 30 & \textbf{0.75} & 0.81 & 1.25 & 0.64 & 0.92 & 2.06 & 0.56 & 0.86 & 2.53 & 0.57 & 0.69 & 1.22 & 0.48 & 0.75 & 1.89 & 0.23 & 0.94 & 5.17\\ & 50 & \textbf{0.92} & 1.0 & 1.19 & 0.83 & 1.0 & 1.39 & 0.61 & 0.86 & 2.14 & 0.61 & 0.69 & 1.42 & 0.55 & 0.81 & 2.14 & 0.28 & 1.0 & 5.08\\ & 70 & \textbf{0.99} & 1.0 & 1.0 & 0.94 & 1.0 & 1.08 & 0.64 & 0.83 & 1.53 & 0.85 & 0.92 & 1.17 & 0.81 & 0.94 & 1.39 & 0.36 & 1.0 & 3.64\\ & 100 & \textbf{1.0} & 1.0 & 1.0 & \textbf{1.0} & 1.0 & 1.0 & 0.67 & 0.75 & 1.17 & \textbf{1.0} & 1.0 & 1.0 & \textbf{1.0} & 1.0 & 1.0 & 0.42 & 1.0 & 2.75\\\hline%
\multirow{5}{*}{ \rotatebox[origin=c]{90}{\textsc{Other}}}%
 & 10 & \textbf{0.78} & 0.89 & 3.27 & \textbf{0.78} & 0.9 & 3.38 & 0.74 & 0.93 & 3.88 & 0.4 & 0.47 & 1.67 & 0.53 & 0.77 & 3.68 & 0.46 & 0.99 & 6.43\\ & 30 & \textbf{0.81} & 0.91 & 1.77 & 0.79 & 0.94 & 2.04 & 0.69 & 0.95 & 2.67 & 0.63 & 0.73 & 1.38 & 0.57 & 0.89 & 2.54 & 0.3 & 0.99 & 5.59\\ & 50 & \textbf{0.88} & 0.95 & 1.32 & 0.84 & 0.97 & 1.63 & 0.77 & 0.95 & 1.86 & 0.77 & 0.85 & 1.16 & 0.7 & 0.93 & 1.84 & 0.29 & 0.98 & 4.8\\ & 70 & \textbf{0.95} & 0.99 & 1.15 & 0.94 & 0.99 & 1.23 & 0.82 & 0.96 & 1.49 & 0.88 & 0.94 & 1.13 & 0.78 & 0.99 & 1.55 & 0.35 & 1.0 & 3.92\\ & 100 & \textbf{0.98} & 1.0 & 1.06 & \textbf{0.98} & 1.0 & 1.06 & 0.9 & 0.97 & 1.24 & 0.95 & 1.0 & 1.08 & 0.87 & 1.0 & 1.29 & 0.45 & 1.0 & 3.01\\\midrule%
\multicolumn{2}{c|}{AVG}  & \textbf{0.86} & 0.94 & 1.79 & 0.84 & 0.95 & 1.99 & 0.77 & 0.95 & 2.39 & 0.7 & 0.79 & 1.31 & 0.67 & 0.91 & 2.22 & 0.39 & 0.99 & 5.2%
\\
\bottomrule
\end{tabular}\\
\vspace{0.2cm}
\begin{tabular}{cc|ccc|ccc|ccc|ccc|ccc|ccc}
\toprule
 \multicolumn{2}{c}{} & \multicolumn{18}{c}{Sub-Optimal}\\
\cmidrule(lr){3-20}
 \multicolumn{2}{c}{} & \multicolumn{3}{c}{\dhc} & \multicolumn{3}{c}{\dhcu} & \multicolumn{3}{c}{\rg} & \multicolumn{3}{c}{\pom} & \multicolumn{3}{c}{\pomA} & \multicolumn{3}{c}{\pomC}\\%
\cmidrule(lr){3-5}\cmidrule(lr){6-8}\cmidrule(lr){9-11}\cmidrule(lr){12-14}\cmidrule(lr){15-17}\cmidrule(lr){18-20}\\
\# & \% %
& AGR & ACC & SPR & AGR & ACC & SPR & AGR & ACC & SPR & AGR & ACC & SPR & AGR & ACC & SPR & AGR & ACC & SPR\\
\midrule
\multirow{5}{*}{ \rotatebox[origin=c]{90}{\textsc{blocks}}}%
 & 10 & 0.41 & 0.86 & 6.86 & 0.42 & 0.89 & 7.42 & \textbf{0.46} & 0.97 & 10.61 & 0.06 & 0.19 & 1.19 & 0.19 & 0.58 & 4.44 & 0.34 & 1.0 & 17.53\\ & 30 & 0.49 & 0.78 & 3.17 & 0.35 & 0.86 & 6.92 & \textbf{0.54} & 1.0 & 4.86 & 0.28 & 0.56 & 1.17 & 0.32 & 0.89 & 3.36 & 0.26 & 1.0 & 13.47\\ & 50 & 0.55 & 0.86 & 3.08 & 0.42 & 0.94 & 5.61 & \textbf{0.62} & 0.97 & 2.72 & 0.39 & 0.72 & 1.08 & 0.36 & 0.81 & 2.17 & 0.27 & 1.0 & 9.89\\ & 70 & \textbf{0.71} & 0.92 & 2.06 & 0.56 & 0.94 & 3.06 & 0.68 & 1.0 & 2.44 & 0.51 & 0.94 & 1.33 & 0.44 & 1.0 & 2.22 & 0.26 & 1.0 & 8.61\\ & 100 & \textbf{0.84} & 1.0 & 1.67 & \textbf{0.84} & 1.0 & 1.67 & 0.8 & 1.0 & 2.08 & 0.51 & 1.0 & 1.67 & 0.48 & 1.0 & 1.92 & 0.28 & 1.0 & 6.42\\\hline%
 \multirow{5}{*}{ \rotatebox[origin=c]{90}{\textsc{ipc-grid}}}%
 & 10 & 0.77 & 0.92 & 1.81 & 0.75 & 0.98 & 2.4 & \textbf{0.81} & 1.0 & 2.73 & 0.64 & 0.88 & 2.23 & 0.59 & 0.98 & 3.44 & 0.42 & 1.0 & 5.6\\ & 30 & 0.82 & 0.94 & 1.13 & 0.77 & 0.98 & 1.6 & \textbf{0.9} & 1.0 & 1.27 & 0.81 & 0.96 & 1.35 & 0.8 & 0.96 & 1.52 & 0.67 & 0.96 & 2.23\\ & 50 & 0.84 & 0.94 & 1.13 & 0.84 & 1.0 & 1.56 & \textbf{0.92} & 1.0 & 1.1 & 0.87 & 1.0 & 1.08 & 0.86 & 1.0 & 1.13 & 0.86 & 1.0 & 1.21\\ & 70 & 0.89 & 1.0 & 1.1 & 0.85 & 1.0 & 1.23 & \textbf{0.93} & 1.0 & 1.02 & 0.92 & 1.0 & 1.0 & 0.92 & 1.0 & 1.0 & 0.92 & 1.0 & 1.0\\ & 100 & \textbf{0.94} & 1.0 & 1.0 & \textbf{0.94} & 1.0 & 1.0 & \textbf{0.94} & 1.0 & 1.0 & \textbf{0.94} & 1.0 & 1.0 & \textbf{0.94} & 1.0 & 1.0 & \textbf{0.94} & 1.0 & 1.0\\\hline%
 \multirow{5}{*}{ \rotatebox[origin=c]{90}{\textsc{sokoban}}}%
 & 10 & \textbf{0.52} & 0.61 & 1.78 & 0.44 & 0.72 & 3.17 & 0.29 & 0.64 & 4.56 & 0.35 & 0.64 & 2.47 & 0.38 & 0.92 & 4.08 & 0.24 & 1.0 & 6.86\\ & 30 & \textbf{0.77} & 0.83 & 1.08 & 0.62 & 0.97 & 2.67 & 0.43 & 0.75 & 2.92 & 0.56 & 0.75 & 1.72 & 0.51 & 0.86 & 2.64 & 0.24 & 0.97 & 5.5\\ & 50 & \textbf{0.79} & 0.92 & 1.17 & 0.66 & 1.0 & 2.58 & 0.53 & 0.72 & 1.83 & 0.58 & 0.75 & 1.39 & 0.53 & 0.86 & 2.31 & 0.25 & 0.97 & 5.14\\ & 70 & 0.8 & 0.97 & 1.03 & \textbf{0.85} & 1.0 & 1.39 & 0.54 & 0.61 & 1.28 & 0.63 & 0.86 & 1.25 & 0.59 & 0.92 & 1.75 & 0.3 & 1.0 & 4.11\\ & 100 & \textbf{0.83} & 1.0 & 1.0 & \textbf{0.83} & 1.0 & 1.0 & 0.58 & 0.58 & 1.33 & \textbf{0.83} & 1.0 & 1.0 & \textbf{0.83} & 1.0 & 1.0 & 0.38 & 1.0 & 2.75\\\hline%
\multirow{5}{*}{ \rotatebox[origin=c]{90}{\textsc{Other}}}%
 & 10 & \textbf{0.72} & 0.89 & 2.84 & 0.7 & 0.92 & 3.29 & 0.63 & 0.94 & 3.81 & 0.45 & 0.56 & 1.69 & 0.48 & 0.82 & 3.63 & 0.37 & 0.99 & 6.35\\ & 30 & \textbf{0.77} & 0.9 & 1.74 & 0.63 & 0.95 & 2.89 & 0.68 & 0.93 & 2.35 & 0.65 & 0.78 & 1.32 & 0.55 & 0.91 & 2.48 & 0.28 & 0.99 & 5.47\\ & 50 & \textbf{0.87} & 0.95 & 1.33 & 0.72 & 0.99 & 2.23 & 0.76 & 0.96 & 1.85 & 0.79 & 0.9 & 1.22 & 0.69 & 0.97 & 1.9 & 0.3 & 1.0 & 4.59\\ & 70 & \textbf{0.94} & 0.98 & 1.16 & 0.83 & 0.99 & 1.63 & 0.84 & 0.96 & 1.45 & 0.87 & 0.95 & 1.1 & 0.79 & 0.98 & 1.52 & 0.37 & 1.0 & 3.81\\ & 100 & \textbf{0.97} & 1.0 & 1.06 & \textbf{0.97} & 1.0 & 1.06 & 0.89 & 0.97 & 1.24 & 0.95 & 1.0 & 1.08 & 0.86 & 1.0 & 1.29 & 0.46 & 1.0 & 3.01\\\midrule%
\multicolumn{2}{c|}{AVG} & \textbf{0.82} & 0.93 & 1.71 & 0.75 & 0.97 & 2.39 & 0.74 & 0.93 & 2.3 & 0.7 & 0.83 & 1.31 & 0.65 & 0.93 & 2.19 & 0.38 & 1.0 & 5.01%
\\
\bottomrule
\end{tabular}\\
\caption{Agreement ratio (AGR), accuracy (ACC) and spread (SPR) for each method on optimal and sub-optimal data sets.}
\label{tab:uncertainty}
\end{table*}

\subsubsection{Evaluating the Constraints }
We measure how the sources of operator-counting constraints impact the quality of the solutions, and if more informed heuristics~$h_{\observations}$ improve the solution of goal recognition tasks.
The constraint sources are: state equation $h^{\text{SEQ}}$~\cite{bonet2013admissible}, landmarks $h^{\text{LMC}}$~\cite{bonet2014flow}, and the post-hoc optimization $h^{\text{PhO}}$~\cite{florian2016posthoc}. 

Figure~\ref{fig-h-value} shows the value of~$\h_{\observations}$ for each source of constraints. 
Each point is the $\h_{\observations}$-value for a goal recognition task with its reference goal in the optimal data set. 
There are four figures in each group, one for each degree of observability (10\%, 30\%, 50\% and 70\%). They show that, in general,~$h^{\text{SEQ}}_{\observations}$ and~$h^{\text{LMC}}_{\observations}$ are more informed than~$h^{\text{PhO}}_{\observations}$, and that~$h^{\text{SEQ}}_{\observations}$ and~$h^{\text{LMC}}_{\observations}$ are comparable. 
Also, as expected, the difference in the values decreases as observability increases. 
On average~$h^{\text{SEQ}}_{\observations}$ and~ $h^{\text{LMC}}_{\observations}$ are more informed than~$h^{\text{PhO}}_{\observations}$ on 61.49\% and 72.58\% of the goal recognition tasks respectively. 
$h^{\text{SEQ}}_{\observations}$ is more informed than~$h^{\text{LMC}}_{\observations}$ on 31.19\% of the tasks, and~$h^{\text{LMC}}_{\observations}$ is more informed than~$h^{\text{SEQ}}_{\observations}$ on 42.96\% of the tasks.

\begin{figure}[tb]
\centering
\hspace{-0.55cm}
\begin{subfigure}{.28\columnwidth}
\centering
\input{scatter2.tex}
\end{subfigure}\quad%
\hspace{0.05cm}%
\begin{subfigure}{.28\columnwidth}
\centering
\input{scatter.tex}
\end{subfigure}\quad%
\hspace{0.1cm}%
\begin{subfigure}{.28\columnwidth}
\centering
\input{scatter3.tex}
\end{subfigure}
\caption{Heuristic values for~$h_{\observations}$.}
\label{fig-h-value}
\end{figure}


\begin{figure}[h]
\centering
\begin{subfigure}{0.3\columnwidth}
\centering
\rotatebox[origin=c]{90}{$\text{AGR}^1 - \text{AGR}^2$}%
\input{quad2.tex}
\end{subfigure}\quad%
\begin{subfigure}{.3\columnwidth}
\centering
\input{quad.tex}
\end{subfigure}\quad%
\begin{subfigure}{.3\columnwidth}
\centering
\input{quad3.tex}
\end{subfigure}
$h^1_{\observations} - h^2_{\observations}$
\caption{Relation between $h_{\observations}$-value and agreement ratio.}
\label{fig-quad}
\end{figure}




Table~\ref{agr-constraints} shows the results of the \textit{agreement} ratio for each source of operator-counting constraints solving goal recognition tasks in the optimal and sub-optimal data sets. The solution set~$\dhc$ is computed using $h_{\observations}$, and when two or more sources of operator-counting constraints are used, they are all combined into a single integer program~$\textup{IP}^{\constraints}_{\observations}$.
The first group of columns shows the results for each source of constraints used individually, and the second combined in pairs. 
When the constraints are used individually $h^{\text{SEQ}}_{\observations}$ and~$h^{\text{LMC}}_{\observations}$ achieve the best results for different domains. 
For example, $h^{\text{SEQ}}_{\observations}$ is the best for \textsc{Blocks} while $h^{\text{LMC}}_{\observations}$ is the best for \textsc{IPC-Grid}. 
When pairs of constraints are combined the results improve and again the pair formed by $h^{\text{LMC}}_{\observations}$ and $h^{\text{SEQ}}_{\observations}$ provides best results. 
Results using all constraints are similar to using the pair $h^{\text{LMC}}_{\observations}$ and $h^{\text{SEQ}}_{\observations}$ (as presented next in Table~\ref{tab:uncertainty}).
There are two key conclusions of these results. 
First, the agreement increases with the degree of observability, but even with 100\% it is still hard to obtain perfect agreement.
Second, the agreement degrades in the sub-optimal data set, but our method maintains an average of $0.82$. 

Figure~\ref{fig-quad} shows the relation between $h_{\observations}$-value and agreement for goal recognition tasks. 
Again, there  are  four  Figures in  each  group,  one for each degree of observability. 
Each point on the $x$-axis shows the difference between the agreement ratio for the solution generated using~$h_{\observations}$ for each goal recognition task in the optimal data set (-1 to 1). The $y$-axis shows the difference of~$h_{\observations}$-values for the same task with its goal reference (-10 to 10). 
For example, in group $h^{\text{LMC}}_{\observations}$ vs. $h^{\text{PhO}}_{\observations}$ the points are clustered on right upper quadrant which shows that in general when AGR$^1$ is higher than AGR$^2$, $h^1_{\observations}$ is also higher than $h^2_{\observations}$. 
We see the same trend in group $h^{\text{SEQ}}_{\observations}$ vs. $h^{\text{PhO}}_{\observations}$. 
We highlight a different situation in the group $h^{\text{SEQ}}_{\observations}$ vs. $h^{\text{LMC}}_{\observations}$ since the points are clustered on the upper right and bottom left quadrants. 
This shows that higher $h_{\observations}$-values tend to produce higher agreement ratio. 
Thus, these results provide evidence that more informed heuristics improve the solution of goal recognition tasks. 

\subsubsection{Evaluating Previous Methods and Uncertainty}

Table~\ref{tab:uncertainty} compares our \dhc{} and \dhcu{} methods to two other polynomial time approaches from the literature, reporting agreement ratio, accuracy and spread for optimal and sub-optimal data sets with degrees of observability with 10\%, 30\%, 50\%, 70\% and 100\% of the observations. 
Our methods use the three sources of constraints $h^{\text{SEQ}}$, $h^{\text{LMC}}$, and $h^{\text{PhO}}$.
\rg~\cite{ramirez2009plan} computes a relaxed plan efficiently and returns as the goal set the goal candidates with relaxed plans that satisfy the largest number of observations. 
\pom~\cite{pereira2017landmark} performs the recognition task by computing landmarks and returns as the goal set the goal candidates that have the highest number of landmarks satisfied by the observation. We use their \emph{goal completion} heuristic for its better results. 
We report the results of \pomA{} and \pomC{}, which return larger goal sets, including those within a 10\% and 30\% threshold of the goals with the highest number of landmarks satisfied. 

On both data sets our approach~\dhc{} has the highest agreement ratio on average and is the best in almost all domains and degrees of observability. 
An exception is the domain \textsc{IPC-Grid} where \rg{} has in general better results. 
We note that in hard domains like \textsc{Sokoban}, our methods have much higher agreements ratios than other approaches. 
For example, on the optimal data set for this domain,~\dhc{} has average agreement ratio of 0.81 while the next best approach \rg{} has average agreement ratio of 0.76. 

\begin{table}[htb]
\centering
    \addtolength{\leftskip} {-6cm} 
    \addtolength{\rightskip}{-6cm}
\fontsize{9.}{9.}\selectfont
\setlength\tabcolsep{1.8pt}
\begin{tabular}{cc|cccc|cccc}
\toprule
 \multicolumn{2}{c}{} & \multicolumn{4}{c}{Optimal} & \multicolumn{4}{c}{Sub-Optimal}\\
\cmidrule(lr){3-6}\cmidrule(lr){7-10}
\# & \% & \dhc & \dhcf & \rg & \pom & \dhc & \dhcf & \rg & \pom\\%
\midrule
\multirow{5}{*}{ \rotatebox[origin=c]{90}{\textsc{blocks}}}%
 & 10 & \textbf{0.32} & \textbf{0.32} & 0.31 & 0.06 & 0.38 & 0.38 & \textbf{0.42} & 0.05\\ & 30 & 0.37 & 0.37 & \textbf{0.39} & 0.13 & 0.36 & 0.37 & \textbf{0.49} & 0.22\\ & 50 & \textbf{0.64} & \textbf{0.64} & 0.6 & 0.37 & 0.53 & 0.53 & \textbf{0.55} & 0.28\\ & 70 & 0.79 & \textbf{0.81} & 0.77 & 0.47 & \textbf{0.67} & \textbf{0.67} & 0.63 & 0.38\\ & 100 & 0.88 & 0.88 & \textbf{0.89} & 0.57 & 0.78 & \textbf{0.82} & 0.74 & 0.51\\\hline%
 \multirow{5}{*}{ \rotatebox[origin=c]{90}{\textsc{ipc-grid}}}%
 & 10 & \textbf{0.57} & \textbf{0.57} & 0.16 & 0.38 & \textbf{0.62} & \textbf{0.62} & 0.12 & 0.54\\ & 30 & \textbf{0.85} & \textbf{0.85} & 0.28 & 0.71 & 0.68 & 0.68 & 0.08 & \textbf{0.72}\\ & 50 & \textbf{0.89} & \textbf{0.89} & 0.07 & 0.81 & 0.84 & 0.84 & 0.04 & \textbf{0.85}\\ & 70 & \textbf{0.95} & \textbf{0.95} & 0.15 & 0.93 & 0.89 & 0.89 & 0.02 & \textbf{0.9}\\ & 100 & \textbf{1.0} & \textbf{1.0} & 0.08 & 0.99 & \textbf{0.94} & \textbf{0.94} & 0.04 & 0.92\\\hline%
 \multirow{5}{*}{ \rotatebox[origin=c]{90}{\textsc{sokoban}}}%
 & 10 & \textbf{0.27} & \textbf{0.27} & 0.1 & 0.24 & 0.31 & \textbf{0.32} & 0.13 & 0.25\\ & 30 & 0.56 & \textbf{0.7} & 0.2 & 0.34 & 0.48 & \textbf{0.56} & 0.12 & 0.29\\ & 50 & 0.61 & \textbf{0.82} & 0.2 & 0.57 & 0.5 & \textbf{0.73} & 0.01 & 0.46\\ & 70 & 0.6 & \textbf{0.97} & 0.08 & 0.84 & 0.54 & \textbf{0.8} & 0.06 & 0.58\\ & 100 & 0.66 & \textbf{1.0} & 0.04 & 0.96 & 0.35 & \textbf{0.85} & 0.04 & 0.77\\\hline%
\multirow{5}{*}{ \rotatebox[origin=c]{90}{\textsc{Other}}}%
 & 10 & \textbf{0.44} & \textbf{0.44} & 0.27 & 0.27 & \textbf{0.42} & \textbf{0.42} & 0.31 & 0.31\\ & 30 & 0.57 & \textbf{0.58} & 0.32 & 0.51 & 0.6 & \textbf{0.63} & 0.36 & 0.53\\ & 50 & 0.77 & \textbf{0.78} & 0.37 & 0.65 & \textbf{0.78} & \textbf{0.78} & 0.41 & 0.71\\ & 70 & 0.88 & \textbf{0.89} & 0.46 & 0.8 & 0.86 & \textbf{0.88} & 0.41 & 0.79\\ & 100 & 0.95 & \textbf{0.96} & 0.45 & 0.9 & 0.91 & \textbf{0.96} & 0.4 & 0.87\\\midrule%

\multicolumn{2}{c|}{AVG} & 0.71 & \textbf{0.73} & 0.35 & 0.61 & 0.69 & \textbf{0.72} & 0.34 & 0.61\\\midrule%
\multicolumn{2}{c|}{ACC} & 0.89 & \textbf{0.9} & 0.5 & 0.73 & 0.87 & \textbf{0.89} & 0.48 & 0.76\\\hline%
\multicolumn{2}{c|}{SPR} & 1.91 & 1.78 & 1.38 & \textbf{1.3} & 1.93 & 1.72 & 1.34 & \textbf{1.31}%
\\
\bottomrule
\end{tabular}\\
\caption{Agreement ratio, and average accuracy (ACC) and spread (SPR) results on data sets with noisy observations.}
\label{tab-noisy}
\end{table}

Table~\ref{tab:uncertainty} also shows accuracy and spread for all methods. 
It shows that many methods can achieve high accuracy while yielding a high spread, thus degrading the agreement ratio. 
For example, while \pomC{} has a perfect accuracy on almost all domains on the sub-optimal data set, its spread is the highest. 
The \textsc{Blocks} domain has on average 20.33 goal candidates, and for \pomC{} to achieve a competitive accuracy on 10\% of observability it returns almost all goals with a spread of 17.53. 
By contrast, our \dhcu{} method increases the accuracy without increasing the spread excessively by measuring uncertainty.
This happens especially in the low observability scenarios it was designed to address. 
Take for example the results of \textsc{Sokoban} on sub-optimal data set, in which \dhcu{} shows a substantially higher accuracy without a corresponding increase in spread, unlike other methods. 
Our idea to measure the uncertainty is general since it does not require linear programming heuristics and could be applied to \rg{} and \pom{} to improve their results. 

\subsubsection{Noisy Observations}

Table~\ref{tab-noisy} compares agreement ratio of our \dhc{} and \dhcf{} methods with \rg{} and \pom{} on noisy data sets. 
The last two rows show the average accuracy and spread over all domains. 
Again, our methods use the three sources of constraints $h^{\text{SEQ}}$, $h^{\text{LMC}}$, and $h^{\text{PhO}}$. 
Here most methods degrade with noisy observations, reducing their agreement ratio. 
\dhcf{}, which addresses noisy observations explicitly, has on average the highest agreement ratio and accuracy on both data sets. 
For example, on the \textsc{Sokoban} domain some noisy observations might be impossible to satisfy because they lead to unsolvable states on all plans. 
In this situation~\dhcf{} substantially improves the agreement ratio.

\section{Discussion}

In this paper we developed a novel class of goal recognition methods based on linear programming models. 
These methods include an uncertainty measurement that increases the accuracy on low observability scenarios, as well as an efficient and automatic method to address noisy observations.
We adapt and provide a benchmark to compare methods using the agreement ratio, which allows us to evaluate our methods in a number of different ways. 
First, we evaluate how different sources of constraints impact the quality of our solutions. 
Second, we assess how our additional constraints and uncertainty measurement affect performance under noise and low observability, respectively. 
Third, we compare our methods to previous ones, showing that ours are, in general, superior.

\section{Acknowledgments}

Felipe Meneguzzi acknowledges support from CNPq with projects 407058/2018-4 (Universal) and 302773/2019-3 (PQ Fellowship). 
Andr\'e G. Pereira acknowledges support from FAPERGS with project 17/2551-0000867-7. This study was financed in part by the Coordena\c c\~ao de Aperfei\c coamento de Pessoal de N\'ivel Superior - Brasil (CAPES) - Finance Code 001. 
Ramon Fraga Pereira acknowledges support from 
the ERC Advanced Grant WhiteMech (No. 834228) and the EU ICT-48 2020 project TAILOR (No. 952215).

\normalsize	
\bibliography{ref} 

\end{document}